%% file: iclr2027_conference.tex
\documentclass{article}
\usepackage{iclr2027_conference,times}

\input{math_commands.tex}

\usepackage{hyperref}
\usepackage{url}

\title{\raisebox{-0.15\height}{
    \includegraphics[height=1.05em]{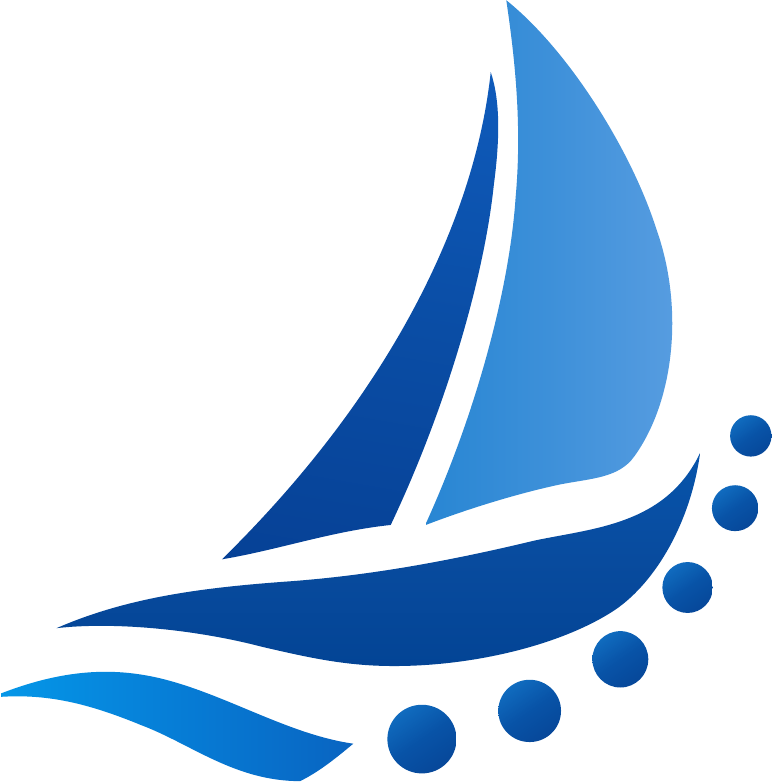}
}
\hspace{0.0em}
Vela: Scaling Vision-Language-Action \\Models with Adaptive Action Curve\\ Parametrization}

\author{Yifan Li$^{1,2}$\thanks{Equal Contribution.} \quad Jiaxu Wang$^{3*}$ \quad Dongming Wu$^{3}$ \quad Yicheng Jiang$^{3}$ \quad Ryan Ji$^{4}$\\ \textbf{Xiangyu Yue$^{3}$\thanks{Corresponding Authors.}  \quad Yanwei Fu$^{1,2\dagger}$}\\
$^{1}$Fudan University \quad $^{2}$Shanghai Innovation Institute\\
$^{3}$The Chinese University of Hong Kong \quad $^{4}$NovaXBot\\
}

\usepackage{booktabs}
\usepackage{xcolor}
\usepackage{colortbl}
\definecolor{gaincolor}{RGB}{34,139,34}
\usepackage{graphicx}
\usepackage{multirow}
\newcommand{\mymodel}[0]{Vela}
\usepackage{pifont}
\newcommand{\cmark}{\ding{51}}
\newcommand{\xmark}{\ding{55}}
\usepackage{amsmath,amssymb}
\usepackage{algorithm}
\usepackage{algpseudocode}
\usepackage{xspace}
\newcommand{\eg}{\emph{e.g.,}\xspace}

\iclrfinalcopy
\begin{document}

\maketitle

\begin{abstract}
Most vision-language-action models represent future motion as fixed-rate action chunks, tying temporal resolution and prediction horizon to a fixed output budget. This pointwise representation wastes capacity on highly correlated neighboring actions, leaves temporal continuity and smoothness to be learned implicitly, and forces a tradeoff between long-horizon coverage and the local precision required for contact-rich manipulation. To address these limitations, we introduce \mymodel{}, a vision-language-action foundation model that represents future robot behavior as continuous trajectories. \mymodel{} combines a compact spline-based action representation with motion-dependent temporal support and a shared action interface for heterogeneous embodiments, allowing a fixed output budget to adapt its temporal resolution across motions.
We pretrain \mymodel{} on large-scale multi-embodiment robot data and evaluate it on LIBERO-X, EBench, and two real-world long-horizon tasks, egg-cake cooking and potato shredding, obtaining promising results across simulation and physical manipulation.
These results highlight the potential of continuous action representations as a foundation for future embodied foundation models.
Project page and more results: \url{https://clementine24.github.io/Vela/}.
\end{abstract}

\section{Introduction}
Vision--Language--Action (VLA) models are rapidly scaling to larger datasets, broader task distributions, and more diverse robot embodiments~\citep{intelligence2026pi,yuan2026qwen,ma2026internvla}. Yet most VLAs still represent future motion as a fixed-length sequence of actions sampled on a predetermined temporal grid. This pointwise interface assigns separate prediction capacity to highly correlated neighboring actions and couples the output budget to both prediction horizon and temporal resolution: representing a longer future requires more action points, while a fixed budget forces coarser sampling. Conversely, spreading the same budget over a long horizon can remove the local precision needed for contact-rich manipulation. Because temporal continuity is learned implicitly, execution also relies on a predefined sampling grid. These limitations reflect a mismatch between current VLA representations and the objects they control: robot actions are not isolated points but samples from continuous motions.

A trajectory-level representation provides a more natural action space. Instead of predicting every temporal sample, a policy can predict a compact set of parameters that defines a continuous trajectory and query that trajectory at the desired temporal locations. This representation compresses correlated actions, exposes temporal continuity, and decouples trajectory geometry from the time span over which it is executed. The same fixed parameter budget can therefore describe a long, smooth motion or a short, finely resolved one, while supporting different control frequencies. Splines provide a practical realization of this idea: a small set of control points specifies a continuous curve that can be evaluated on arbitrary temporal grids.

Trajectory-level representations have a long history in robotics, including dynamical and probabilistic movement primitives~\citep{ijspeert2013dynamical,paraschos2013probabilistic}. Recent learning-based policies have explored dynamical-system parameterizations~\citep{bahl2020ndp}, differentiable trajectory optimization~\citep{wan2024difftori}, and geometric curve representations~\citep{li2026gpi}. Spline-based policies have further studied compact action tokenization, asynchronous execution, and temporal rescaling~\citep{zhou2026beast,yang2026abpolicy,han2026b}. These studies show the value of structured motion representations, but they mainly focus on task-specific policies, individual robot platforms, or downstream action tokenization and execution. Whether trajectory-level action modeling can serve as the output space for foundation-scale VLA pretraining across large, heterogeneous robot datasets remains largely unexplored.

To address this gap, we build a VLA that learns future motion in trajectory space and make this representation trainable at foundation scale. A shared cross-embodiment action interface and scalable supervision procedure allow the same model to absorb heterogeneous robot corpora. We further make the temporal span of each prediction motion-dependent: with a fixed number of spline control points, each demonstrated motion is assigned a sample-specific horizon by selecting the longest span that adequately represents it under a noise-aware criterion. Smooth motions can therefore use longer spans, whereas motions with rapid local variation can use shorter spans without changing the output budget. We train this framework on large multi-embodiment robot data and evaluate its transfer to downstream manipulation tasks.

Our main contributions are:
\begin{itemize}
    \item We show how trajectory-space action modeling can be scaled to VLA pretraining through a shared cross-embodiment action interface and a training procedure designed for large heterogeneous robot corpora.
    \item We introduce motion-dependent horizon adaptation, selecting for each demonstrated motion the longest temporal span that a fixed spline budget can represent under a noise-aware criterion.
    \item We evaluate the framework with controlled comparisons, downstream manipulation benchmarks, and ablations, showing improvements from trajectory-space modeling, large-scale pretraining, and horizon adaptation, particularly under contact-rich, long-horizon, and distribution-shifted tasks.
\end{itemize}

\section{Related Work}
\noindent\textbf{Vision-Language-Action Foundation Models.}
Extensive research on vision-language models has advanced cross-modal alignment and downstream visual reasoning~\citep{li2024unified, li2025revseg}.
Building on this progress, vision-language-action (VLA) foundation models scale robot learning across heterogeneous tasks, data, and embodiments by coupling pretrained vision-language representations~\citep{beyer2024paligemma,bai2025qwen3} with generalist action policies~\citep{team2024octo,black2024pi_0,bjorck2025gr00t}. Recent systems further improve cross-task and cross-embodiment generalization through heterogeneous co-training, embodiment alignment, and predictive supervision~\citep{intelligence2025pi_, intelligence2026pi, yuan2026qwen, zheng2026x, cai2026internvla, ma2026internvla, zhou2026tau0wm, wang2026mvista4d}; related robot-control work also studies structured visual representations~\citep{jiang2026learning,wang2024gsrl}. Yet the low-level motion interface of modern chunk-based VLAs remains largely organized around fixed-resolution sequences of sampled future actions.
Recent work begins to relax this temporal rigidity: MoH~\citep{jing2025mixture} models multiple prediction horizons, while AutoHorizon~\citep{wang2026vla} and AAC~\citep{liang2026adaptive} use attention patterns and action entropy to determine execution lengths. These methods still operate on sampled action sequences.
In contrast, \mymodel{} makes temporal support part of the action representation itself, using a fixed set of control points to parameterize continuous trajectories with motion-adaptive future spans.

\noindent\textbf{Spline-Based Robot Policies.} Structured representations of robot motion have long been studied through dynamical and probabilistic movement primitives~\citep{ijspeert2013dynamical,paraschos2013probabilistic}. Recent work incorporates splines into learned policies for efficient action tokenization ~\citep{zhou2026beast}, asynchronous execution~\cite{yang2026abpolicy}, and structured action generation across policy backbones~\cite{tian2026spline}. B-spline Policy~\cite{han2026b} accelerates visuomotor policies through adaptive spline fitting and temporal rescaling. These works primarily study splines as policy-level representations, tokenizers, or execution interfaces. To our knowledge, \mymodel{} is the first VLA foundation model established in trajectory space through large-scale pretraining on multi-embodiment robot data. Our framework is designed to support this scaling: a unified action interface accommodates heterogeneous embodiments, while a noise-calibrated horizon procedure constructs spline targets on the fly. Together, \mymodel{} constructs consistent trajectory targets from heterogeneous datasets, investigates spline trajectories as the output space of a foundation model.

\section{Methodology}\label{sec:methodology}
\begin{figure}
    \centering
    \includegraphics[width=\linewidth]{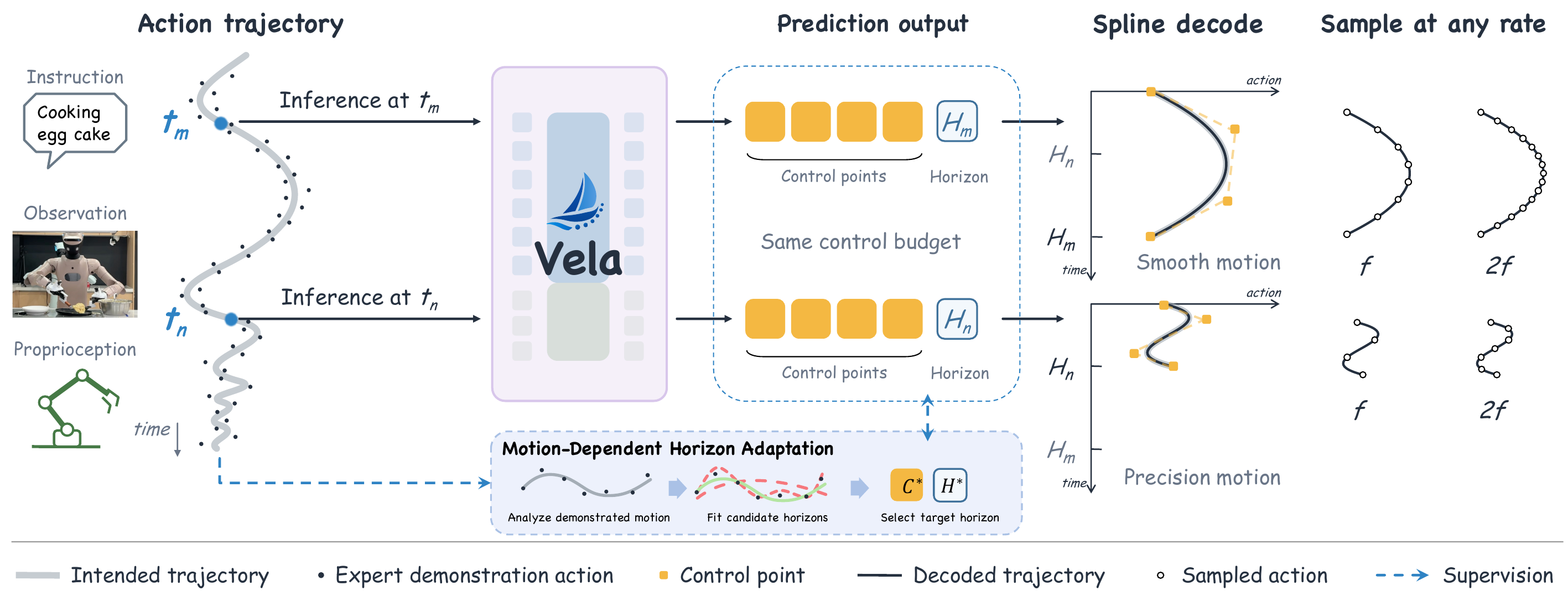}
    \vspace{-5mm}
    \caption{Overview of \mymodel{}. At inference, \mymodel{} jointly predicts a fixed set of spline control points and a motion-dependent horizon, defining a continuous action trajectory that can be sampled at arbitrary control rates.
    During training, \emph{motion-dependent horizon adaptation} constructs supervision targets that adequately represent the demonstrated motion.
    The same control point budget thus covers longer spans for smooth motion and shorter spans when finer temporal resolution is needed.}
    \label{fig:method}
\end{figure}
We develop \mymodel{} by treating demonstrated actions as samples of a continuous trajectory and learning to predict a compact representation of that trajectory. Building on this formulation, we describe the policy, construct motion-dependent horizons, and scale training to heterogeneous robot data.

\subsection{Preliminaries: What should a VLA predict?}\label{sec:preliminaries}

\noindent\textbf{Action prediction.}
Conventional VLA policies represent future motion as an action
chunk conditioned on the current observation and instruction. At timestep $t$, let $z_t=(\mathbf{v_t},\mathbf{s_t},\ell)$ denote the multimodal policy context, consisting of visual observations $\mathbf{v_t}$, proprioceptive state $\mathbf{s_t}$, and a language instruction $\ell$. A conventional Vision-Language-Action (VLA) policy parameterizes a conditional distribution over a future action chunk,
\begin{equation}
    \mathbf A_{t,H} \sim \pi_\theta(\cdot\mid z_t) \qquad \mathbf{A_{t,H}} = [\mathbf{a_t}, \mathbf{a_{t+1}}, \ldots, \mathbf{a_{t+H-1}}]^\top\in\mathbb R^{H\times d_a},
    \label{eq:vla_chunk}
\end{equation}
where $H$ is the prediction chunk horizon and $d_a$ is the action dimension. Regardless of the generative objective, this representation describes motion through its values on a discrete temporal grid.

\noindent\textbf{From actions to trajectories.} We instead view a demonstrated action chunk as a sampled realization of an underlying continuous motion. For a given future window, let $\gamma_t:[0,1]\to\mathbb R^{d_a}$ denote its \textit{latent} intended trajectory, and write the recorded actions as
\begin{equation}
    \mathbf a_{t+i}
    =
    \boldsymbol{\gamma}_t(u_i)
    +
    \boldsymbol{\epsilon}_{t,i},
    \qquad u_i=\frac{i}{H-1},
    \quad i=0,\ldots,H-1,
    \label{eq:trajectory_sampling}
\end{equation}
where $\boldsymbol{\epsilon}_{t,i}$ captures deviations introduced during demonstration and execution. This view separates the motion to be represented from a particular noisy recording of it. This modeling view motivates a structured approximation to the underlying motion rather than adopting its sampled realization as the action representation.

To obtain a finite-dimensional representation~\citep{de1968uniform, de1978practical}, we approximate the path $\boldsymbol{\gamma}_t$ using a B-spline curve,
\begin{equation}
    \hat\gamma_t(u)
    =\sum_{n=0}^{N-1}b_n(u)\,\mathbf c_{t,n},
    \qquad u\in[0,1],
    \label{eq:spline_curve}
\end{equation}
where $\mathbf{b}_{n}$ are cubic B-spline basis functions with uniform knots, and $\mathbf c_{t,n}\in\mathbb R^{d_a}$ are control points. The resulting curve is determined by
$\mathbf C_t=[\mathbf c_{t,0},\ldots,\mathbf c_{t,N-1}]^\top
\in\mathbb R^{N\times d_a}$.
The continuous motion coordinates are \(C^2\)-continuous at simple interior knots within each predicted segment and can be queried at arbitrary temporal locations.
The name \mymodel{}, inspired by sailing, reflects this view of future motion as a continuous course described by a compact set of parameters.

\subsection{Vela Model Architecture}\label{sec:spline_vla}
\noindent\textbf{Policy backbone.} \mymodel\ follows the $\pi_{0.5}$~\citep{intelligence2025pi_} architecture, consisting of a multimodal vision-language backbone and a flow-matching action expert.
The backbone encodes the policy context as $\mathbf g=G_\phi(z)$, and the action expert $F_\theta$ defines the conditional flow
\begin{equation}
    \frac{\mathrm d\mathbf y^\tau}{\mathrm d\tau}
    =F_\theta(\mathbf y^\tau,\tau;\mathbf g),
    \qquad
    \mathbf y^0\sim\mathcal N(\mathbf 0,\mathbf I),
    \quad \tau\in[0,1],
    \label{eq:vela_flow}
\end{equation}
whose terminal state specifies the predicted motion. Here, $\tau$ denotes generative flow time.

\noindent\textbf{Trajectory prediction.} Rather than representing future motion by samples on a fixed temporal grid, \mymodel\ parameterizes each future motion by two complementary quantities: its geometry and its temporal span. The geometry is parameterized by a set of $N$ control points $\mathbf{C}$, which specify the curve, while a scalar horizon $H$ specifies its extent in action steps. We model them jointly as $\mathbf y=(\mathbf C,H)$ and predict them from $\mathbf y^1$. In implementation, $H$ occupies an additional scalar
output channel alongside the control-point coordinates. Thus, the number of predicted control points remains fixed, while the span of future motion they describe can vary with the context.

\noindent\textbf{Action decoding.}
The horizon maps the normalized curve to the future action timeline:
$\mathbf a(s)=\hat\gamma(s/(H-1))$, $s\in[0,H-1]$.
Evaluating at integer offsets gives
\begin{equation}
    \hat{\mathbf A}_H=\mathbf B_H\mathbf C,
    \qquad
    [\mathbf B_H]_{i,n}
    =b_n\!\left(\frac{i}{H-1}\right),
    \label{eq:trajectory_query}
\end{equation}
where $\mathbf B_H\in\mathbb R^{H\times N}$.
Other query grids change the sampling density without changing the predicted representation.
The remaining question is how to choose a suitable horizon for each demonstrated motion under the fixed control-point budget.

\subsection{Motion-Dependent Horizon Adaptation}
\label{sec:adaptive horizon}

With $N$ fixed, different temporal spans induce different allocations of the same representational capacity. Extending the horizon $H$ distributes the same curve parameters over more action steps. This suits smooth, directionally consistent behaviors such as reaching and transport. Conversely, contact, alignment, and rapid corrective motion may require concentrating the parameters over a shorter interval to preserve local temporal variation. Existing spline-based policies typically predict over a globally fixed temporal span~\citep{zhou2026beast,tian2026spline,yang2026abpolicy}, imposing a single compromise between temporal coherence and local expressivity across all motions.

We instead formulate this trade-off as a horizon-selection problem and seek the longest temporal span that remains sufficient for the demonstrated motion. Let $\mathcal H$ denote the candidate horizons, each satisfying $H>N$:
\begin{equation}
    H^\star
    =\max\{H\in\mathcal H:\psi(H)\le\kappa\},
    \label{eq:temporal_allocation}
\end{equation}
where $\psi(H)$ measures whether the demonstrated motion is adequately represented at span $H$ and $\kappa$ is a shared tolerance.
The central problem is therefore to construct $\psi(H)$ such that insufficient temporal resolution can be distinguished from variation already inherent in the demonstrations.

\noindent\textbf{Noise-aware horizon selection.}
Here, noise refers to variation introduced by teleoperation, sensing, and execution rather than the underlying motion. We estimate a per-source, per-coordinate reference scale from normalized demonstrations and use it to distinguish unresolved motion structure from variation that is already present in the demonstrations.
We assess each candidate horizon by fitting its demonstrated action sequence with the spline basis from \eqref{eq:trajectory_query}: \begin{equation}
    \mathbf C_H=\mathbf B_H^\dagger\mathbf A_H,
    \label{eq:spline_fitting}
\end{equation}
where $\dagger$ denotes the Moore--Penrose pseudoinverse. As suggested by \eqref{eq:trajectory_sampling}, the fitting residual $r(H)=\mathbf A_H-\mathbf B_H\mathbf C_H$ contains both motion not captured by the spline and demonstration variation. Its raw magnitude alone therefore does not indicate whether the temporal span is too long.

To obtain a reference scale, consider a continuous action coordinate $j$ and assume that its demonstration perturbations $E[\epsilon^{(j)}]=0$ and $\mathrm{Cov}[\epsilon^{(j)}]=\sigma_j^2\mathbf I$. For a fixed underlying trajectory, let $\boldsymbol\gamma_j$ denote its samples over the candidate window. The expected residual decomposes as
\begin{equation}
    \mathbb E[\|\mathbf r_j(H)\|_2^2]
    =
        \|(\mathbf I-\mathbf B_H\mathbf B_H^\dagger)
        \boldsymbol\gamma_j\|_2^2
    +
        (H-N)\sigma_j^2,
    \label{eq:residual_decomposition}
\end{equation}
which consists of the first term, unrepresented motion and the second term, demonstration variation. Thus, even when the underlying motion is fully representable, the residual is nonzero and scales with both the noise variance and the residual degrees of freedom $H-N$. A raw-error threshold would mix these effects with the approximation error that horizon selection is intended to control.

We therefore normalize the residual by its noise reference:
\begin{equation}
    \psi^2(H)
    =\frac{1}{|\mathcal J|(H-N)}
      \sum_{j\in\mathcal J}
      \frac{\|\mathbf r_j(H)\|_2^2}{\hat\sigma_j^2},
    \label{eq:scale_adequacy}
\end{equation}
where $\mathcal J$ contains the continuous action dimensions and $\widehat\sigma_j$ is estimated from temporal second differences of normalized demonstrations, separately for each data source. Under the reference noise model and accurate scale estimates, the expectation of $\psi^2(H)$ is approximately one for
representable motion and increases with unresolved structure. This calibration reduces the tendency to shorten a horizon merely to fit demonstration fluctuations.

Gripper commands are evaluated separately because their sparse transitions are not appropriately assessed by division by a near-zero noise scale.
Including such channels in the noise-aware statistic \eqref{eq:scale_adequacy} would therefore be both numerically ill-conditioned and semantically inappropriate. Instead, for each such valid channel \(j\notin\mathcal J\), we reject candidate spans whose normalized projection error  $\|\mathbf r_j(H)\|_2/\sqrt H$ exceeds a small tolerance $\eta$.

\noindent\textbf{On-the-fly target construction.}
The selected fit yields the supervision target $\mathbf y^\star=(\mathbf C^\star,H^\star)$, with $\mathbf C^\star=\mathbf C_{H^\star}$.
Candidate fits and residuals are independently evaluated using batched matrix operations for each sampled transition.
This avoids a separate trajectory annotation pass or iterative knot optimization, and supports shuffled training over large heterogeneous robot corpora. At deployment, no fitting or horizon search is performed:
The policy predicts both quantities from the current context.

\subsection{Trajectory-Space Pretraining and Objectives}\label{sec:mid_training}
We extend the target construction above to large-scale trajectory-space pretraining through a shared action representation and flow-matching objective. The action interface accommodates different embodiments,
while source-specific normalization accounts for differences
in motion statistics.

\noindent\textbf{Shared action space.} To scale training across heterogeneous robot datasets, we map all data sources to a shared $d_a=26$ dimensional action-path space. Each arm contributes $10$ dimensions describing end-effector motion, including $3$-D translation, $6$-D rotation, and a $1$-D gripper command. The remaining $6$ dimensions describe the head, waist, and base motion with two coordinates each.
Per-dimension validity masks exclude unavailable coordinates from target selection and supervision.

Each source uses its own per-coordinate affine quantile normalization and estimates its noise scales $\hat{\sigma}_j$ in the same normalized coordinates. This allows the shared tolerance $\kappa$ to act on a dimensionless, noise-normalized statistic rather than on source-dependent raw errors. Because spline fitting acts along time and independently across action coordinates, the same fixed bases and horizon adaptation procedure apply throughout the training mixture.

\noindent\textbf{Training Objective.}
Given a target $\mathbf y^\star=(\mathbf C^\star,H^\star)$, Gaussian noise $\mathbf y^0$, and flow time $\tau$, the interpolated input is $\mathbf y^\tau=(1-\tau)\mathbf y^0+\tau\mathbf y^\star$. We write the control-point and horizon components of the flow prediction error as
\begin{equation}
    (\mathbf E_C,\mathbf E_H)
    =F_\theta(\mathbf y^\tau,\tau;\mathbf g)
     -(\mathbf y^\star-\mathbf y^0).
    \label{eq:flow_prediction_error}
\end{equation}
For the trajectory geometry, a direct choice is to apply the standard flow-matching objective to the control point coordinates (CP-FM): $\mathcal L_{\mathrm{CP}}=\mathbb E[\|\mathbf E_C\|_F^2/(N d_a)]$. This objective weights prediction errors isotropically in the control point space, without accounting for their different effect on the decoded trajectory, which depend on the spline basis. We instead use decoded-trajectory flow matching (DT-FM), which evaluates the same error after applying the linear decoder:
\begin{equation}
    \mathcal L_{\mathrm{DT}}
    = \mathbb E[
       \frac{\|\mathbf B_{H^\star}\mathbf E_C\|_F^2}
            {H^\star d_a}].
    \label{eq:dt_fm}
\end{equation}

Equivalently, DT-FM replaces the isotropic weighting of CP-FM with the decoder-induced quadratic weighting $\mathbf B_H^\top\mathbf B_H/H$, thereby emphasizing control-point errors according to their effect on the decoded trajectory.
DT-FM is our default geometry objective. Its relation to CP-FM and the effect of this weighting are examined in Appendix~\ref{app:objective} and Sec.~\ref{sec:ablation_study}.

The temporal-support component $H$ is trained with the standard flow-matching objective, denoted by $\mathcal L_H=\mathbb E[\|E_H\|^2]$.
Our final training objective is
\begin{equation}
    \mathcal L
    =
    \mathcal L_{\mathrm{DT}}
    +
    \lambda_H\mathcal L_H .
    \label{eq:final_objective}
\end{equation}

\section{Experiments}
\subsection{Experimental Setup}\label{sec:exp_setup}
\noindent\textbf{Implementation Details.}
\mymodel\ follows the $\pi_{0.5}$~\citep{intelligence2025pi_} architecture, comprising a 3B PaliGemma~\citep{beyer2024paligemma} vision-language backbone and a 300M flow-matching action expert, without introducing additional trainable parameters.
Our study examines trajectory-space action modeling at scale without relying on a larger network or benchmark-specific input.
Starting from pretrained PaliGemma weights, we pretrain \mymodel{} directly in
trajectory space on approximately 20,000 hours of public robot data from 60 datasets spanning 11 embodiment classes, together with approximately 20,000 hours of private robot data.
For comparison, we also introduce $\pi_{0.5}$-Vela, which is initialized from the same $\pi_{0.5}$ base checkpoint as the point-wise baseline. It applies our trajectory representation and supervision during benchmark-specific post-training, without undergoing our trajectory-space pretraining stage.

We map each data source to the shared 26-D action-path interface described in Sec.~\ref{sec:mid_training}, with source-specific normalization and per-dimension validity masks.
We use cubic B-splines with a fixed budget of $N=12$ control points and candidate horizons $\mathcal H=\{16,17,\ldots,40\}$.
At inference time, the conditional flow is integrated using 10 Euler steps.
Additional training and implementation details are provided in Appendix.~\ref{app:pretraining_details}.

\noindent\textbf{Benchmarks and Evaluation.}
We evaluate \mymodel{} on two complementary simulation benchmarks, LIBERO-X and EBench, and through real-world experiments on a wheeled dual-arm robot with 7-DoF arms and three cameras mounted on the head and wrists.
The real-world study considers two long-horizon tasks that require bimanual coordination, tool use, and contact-rich manipulation.
We report the egg-cake cooking task and potato-shredding task in Sec.~\ref{sec:real_world} and Appendix~\ref{app:additional_real_world}, respectively.
\textbf{LIBERO-X}~\citep{wang2026libero}
is a MuJoCo-based tabletop manipulation benchmark evaluating generalization across five levels of progressively stronger spatial, visual, and instruction perturbations. We report success rate (SR) at each level and overall.
\textbf{EBench}~\citep{gao2026ebench} is an Isaac Sim benchmark for generalist dual-arm mobile manipulation, comprising 26 tasks across three complementary regimes.
Its evaluation covers background, object, instruction, and mixed distribution shifts. We report SR and the official task-progress Score for each task family and overall.

For each simulation benchmark, we follow its official $\pi_{0.5}$ fine-tuning recipe and evaluation protocol. Further training and execution details can be found in \ref{app:benchmark_finetuning_details}.

\subsection{Simulation Evaluation}
\noindent\textbf{Simulation Evaluation on LIBERO-X.}
LIBERO-X evaluates robust manipulation under progressively stronger distribution shifts. As shown in Tab~\ref{tab:main_results_liberox}, \mymodel{} achieves the highest average performance among the compared methods. Its advantage over $\pi_{0.5}$ is more pronounced at levels 3 to 5 than at levels 1 and 2, indicating that the gains extend to stronger distribution shifts.

Introducing trajectory-space modeling during benchmark post-training
also improves performance. Starting from the same base checkpoint and using the same fine-tuning recipe as $\pi_{0.5}$, $\pi_{0.5}$-Vela improves SR at every level and reaches $42.0\%$ on average, compared with $39.3\%$ for the point-wise baseline. The results support trajectory-space action modeling for both adapting an existing VLA and pretraining a robot policy over heterogeneous robot data.

\begin{table}
    \centering
    \vspace{-5pt}
    \caption{Evaluation on LIBERO-X~\citep{wang2026libero} across five progressively challenging levels.
    We report success rate (\%), with \textcolor{gaincolor}{green} values indicating absolute gains over baseline $\pi_{0.5}$. $\pi_{0.5}$-\mymodel{} denotes variant initialized from $\pi_{0.5}$ and post-trained in trajectory space.
    }
    \label{tab:main_results_liberox}
    \resizebox{\linewidth}{!}{
    \setlength{\tabcolsep}{7pt}
    \renewcommand{\arraystretch}{1.10}
    \begin{tabular}{lcccccc}
        \toprule
        \textbf{Model}
        & \textbf{Level 1}
        & \textbf{Level 2}
        & \textbf{Level 3}
        & \textbf{Level 4}
        & \textbf{Level 5}
        & \textbf{Avg.} \\
        \midrule

        OpenVLA-OFT~\citep{kim2025fine}
        & 29.0 & 17.6 & 8.8 & 6.4 & 4.2 & 13.2 \\

        $\pi_{0}$~\citep{black2024pi_0}
        & 29.4 & 21.9 & 11.0 & 7.6 & 5.1 & 15.0 \\

        X-VLA~\citep{zheng2026x}
        & 30.1 & 22.6 & 10.3 & 6.0 & 4.1 & 14.6 \\

        OpenWAM-$\alpha$~\citep{wang2026openwam}
        & 35.9 & 28.2 & 21.6 & 13.0 & 10.0 & 21.7 \\

        GR00T N1.5~\citep{bjorck2025gr00t}
        & 43.3 & 32.9 & 18.7 & 13.3 & 9.7 & 23.6 \\

        ${\tau_0}$-VLA~\citep{cai2026tau_0} & 46.5 & 34.2 & 20.5 & 13.0 & 10.5 & 24.9 \\

        Xiaomi-R0~\citep{cai2026xiaomi}
        & \textbf{71.4} & 49.6 & 33.7 & 22.1 & 17.2 & 38.8\\

        $\pi_{0.5}$~\citep{intelligence2025pi_}
        & 65.2 & 53.2 & 36.0 & 24.1 & 18.0 & 39.3 \\

        \midrule
        $\pi_{0.5}$-\mymodel
        & 66.1 & \underline{53.5} & \underline{40.7} & \underline{27.3} & \underline{22.5} & \underline{42.0} \\
        \textbf{\mymodel}
        & \underline{67.2}\,{\scriptsize\textcolor{gaincolor}{(+2.0)}}
        & \textbf{55.9}\,{\scriptsize\textcolor{gaincolor}{(+2.7)}}
        & \textbf{44.9}\,{\scriptsize\textcolor{gaincolor}{(+8.9)}}
        & \textbf{31.5}\,{\scriptsize\textcolor{gaincolor}{(+7.4)}}
        & \textbf{27.1}\,{\scriptsize\textcolor{gaincolor}{(+9.1)}}
        & \textbf{45.3}\,{\scriptsize\textcolor{gaincolor}{(+6.0)}} \\
        \bottomrule
    \end{tabular}}
\end{table}

\noindent\textbf{Simulation Evaluation on EBench.}
EBench evaluates generalist manipulation under diverse control demands, spanning \textit{dexterous-and-precise tabletop interaction} involving sub-centimeter insertion, alignment, and bimanual coordination, \textit{mobile pick-and-place} (PnP), and \textit{long-horizon mobile} tasks composed of extended multi-stage sequences. As shown in Tab.~\ref{tab:main_results_ebench}, \mymodel{} achieves the highest overall SR and task-progress Score among the compared methods, and leads both metrics on PnP and Long-Horizon tasks. Relative to $\pi_{0.5}$, its largest improvement occurs on dexterous-and-precise tabletop manipulation, which demands fine-scale motion representation.

Trajectory space fine-tuning variant $\pi_{0.5}$-Vela also improves overall performance to $42.0\%$ SR and a Score of $58$, with family-wise improvements on both Table Top and PnP tasks. Additionally, \mymodel{}, pretrained directly in trajectory space, outperforms both $\pi_{0.5}$ and
$\pi_{0.5}$-Vela on all metrics.
These results demonstrate the breadth of the pretrained trajectory-space policy across precision-intensive interaction and extended execution, rather than an advantage confined to a single manipulation regime. Further result details are provided in Appendix~\ref{app:further_exp}.

\begin{table}
    \centering
    \vspace{-5pt}
    \caption{Evaluation on EBench~\citep{gao2026ebench} across its three manipulation task families. We report success rate (SR, \%) and the official task-progress Score (Score, \%) with \textcolor{gaincolor}{green} values indicating the absolute gains over baseline $\pi_{0.5}$.}
    \label{tab:main_results_ebench}
    \resizebox{\linewidth}{!}{
    \setlength{\tabcolsep}{5.5pt}
    \renewcommand{\arraystretch}{1.10}
    \begin{tabular}{lcccccccc}
        \toprule
        \multirow{2}{*}{\textbf{Model}}
        & \multicolumn{2}{c}{\textbf{Table Top}}
        & \multicolumn{2}{c}{\textbf{Simple PnP}}
        & \multicolumn{2}{c}{\textbf{Long Horizon}}
        & \multicolumn{2}{c}{\textbf{Overall}} \\
        \cmidrule(lr){2-3}
        \cmidrule(lr){4-5}
        \cmidrule(lr){6-7}
        \cmidrule(lr){8-9}
        & \textbf{SR}
        & \textbf{Score}
        & \textbf{SR}
        & \textbf{Score}
        & \textbf{SR}
        & \textbf{Score}
        & \textbf{SR}
        & \textbf{Score} \\
        \midrule

        X-VLA~{\scriptsize\citep{zheng2026x}}
        & 8.6 & 24
        & 50.0 & 54
        & 6.2 & 25
        & 23.7 & 36 \\

        InternVLA-A1~{\scriptsize\citep{cai2026internvla}}
        & 4.3 & 11
        & 43.0 & 47
        & 17.9 & 46
        & 23.9 & 36 \\

        FastWAM~{\scriptsize\citep{yuan2026fast}}
        & 2.9 & 13
        & 49.5 & 53
        & 16.9 & 38
        & 25.6 & 37 \\

        Cosmos3-Edge~{\scriptsize\citep{agarwal2026cosmos}}
        & 18.6 & 32
        & 41.5 & 44
        & 24.1 & 49
        & 29.3 & 42 \\

        GigaBrain-0.7~{\scriptsize\citep{team2026gigabrain}}
        & 37.9 & 59
        & 42.0 & 45
        & 20.0 & 37
        & 33.3 & 46 \\

        $\pi_{0}$~{\scriptsize\citep{black2024pi_0}}
        & 32.9 & 49
        & 41.5 & 45
        & 25.7 & 49
        & 33.7 & 47 \\

        InternVLA-A1.5~{\scriptsize\citep{ma2026internvla}}
        & 12.1 & 22
        & 50.0 & 53
        & 33.7 & \underline{59}
        & 34.2 & 46 \\

        Qwen-RobotManip~{\scriptsize\citep{yuan2026qwen}}
        & \textbf{50.0} & \textbf{70}
        & 56.5 & 60
        & 29.9 & 55
        & \underline{45.6} & \underline{60} \\

        $\pi_{0.5}$~{\scriptsize\citep{intelligence2025pi_}}
        & 29.3 & 45
        & 56.5 & 61
        & \underline{34.1} & 55
        & 41.4 & 54 \\

        \midrule
        $\pi_{0.5}$-\mymodel{}
        & 34.3 & 54
        & \underline{59.0} & \underline{63}
        & 28.5 & 55
        & 42.0 & 58 \\

        \textbf{\mymodel}
        & \underline{44.3}\,{\scriptsize\textcolor{gaincolor}{(+15.0)}} & \underline{64}\,{\scriptsize\textcolor{gaincolor}{(+19)}}
        & \textbf{63.0}\,{\scriptsize\textcolor{gaincolor}{(+6.5)}} & \textbf{67}\,{\scriptsize\textcolor{gaincolor}{(+6)}}
        & \textbf{39.1}\,{\scriptsize\textcolor{gaincolor}{(+5.0)}} & \textbf{65}\,{\scriptsize\textcolor{gaincolor}{(+10)}}
        & \textbf{49.7}\,{\scriptsize\textcolor{gaincolor}{(+8.3)}} & \textbf{66}\,{\scriptsize\textcolor{gaincolor}{(+12)}} \\

        \bottomrule
    \end{tabular}}
\end{table}
\vspace{-2mm}

\subsection{Real-World Evaluation}\label{sec:real_world}
\begin{figure}
    \centering
    \includegraphics[width=\linewidth]{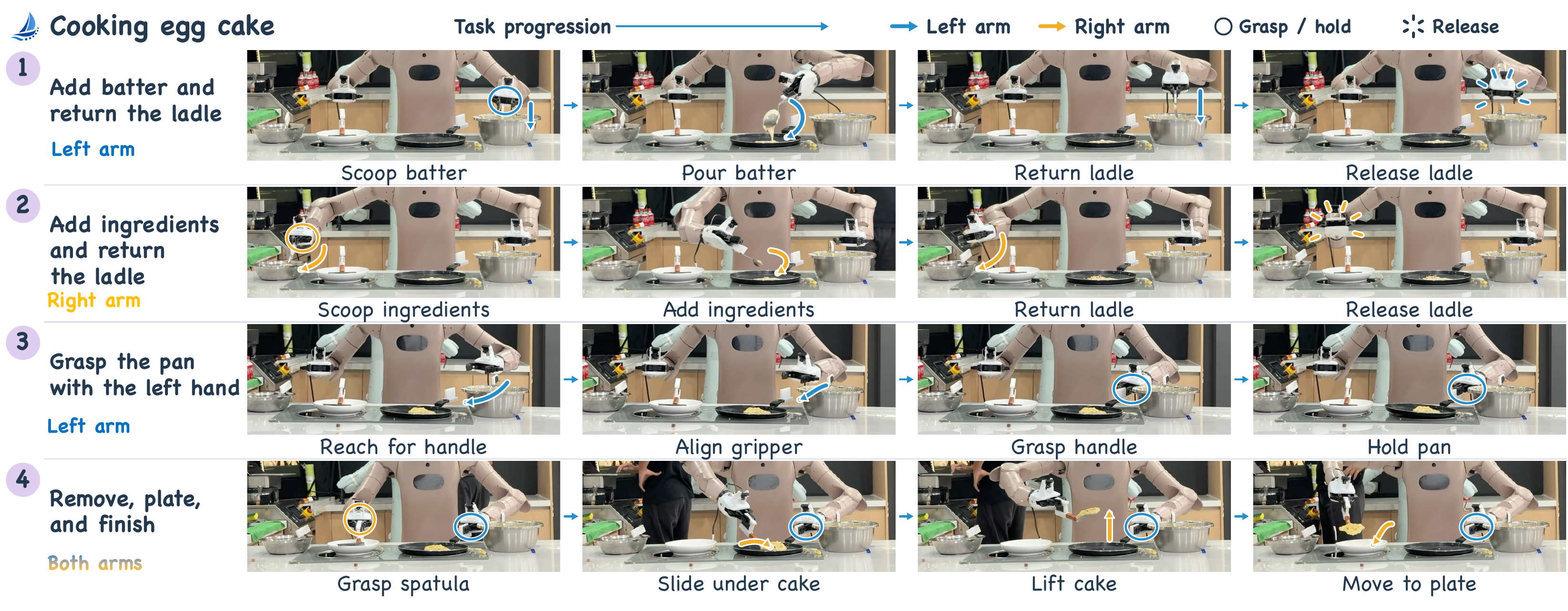}
    \vspace{-15pt}
    \caption{\mymodel{} executing \emph{cooking egg cake}, which spans four subtasks (top to bottom).}
    \label{fig:demo_egg_cake}
\end{figure}

\begin{figure}
    \centering
    \includegraphics[width=0.9\linewidth]{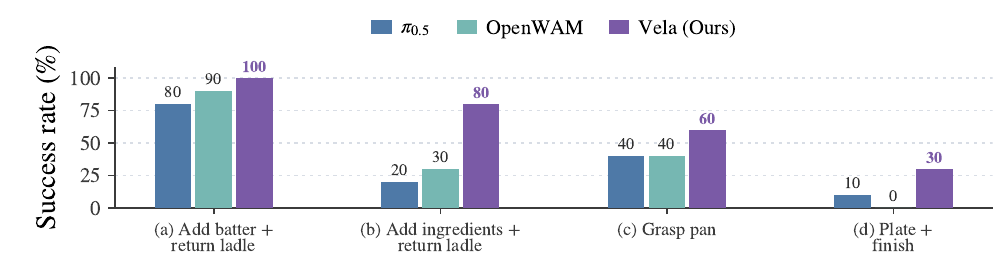}
    \vspace{-15pt}
    \caption{Real-world results on the \emph{cooking egg cake} task across four subtasks.}
    \label{fig:egg_cake_evaluation}
    \vspace{-2mm}
\end{figure}

We further evaluate \mymodel{} on a real-world multi-stage \emph{cooking egg cake} task using the wheeled dual-arm robot. The workspace contains a cooking pan, a batter ladle, ingredient containers, a spatula, and a serving plate. As illustrated in Fig.~\ref{fig:demo_egg_cake}, the task comprises four ordered stages:
\begin{enumerate}
    \item \textbf{Add batter and return the ladle:} Fill the ladle with batter, pour it into the pan, and return the ladle.
    \item \textbf{Add ingredients and return the ladle:} pick up the ingredients, add them to the egg cake, and return the ladle.
    \item \textbf{Grasp the pan with the left hand:} move the left arm to the pan and grasp it firmly.
    \item \textbf{Remove, plate, and finish:} pick up the spatula with the right hand, remove the egg cake from the pan, place it on the plate, and complete the finishing actions.
\end{enumerate}
Each subtask is evaluated over 10 trials. Figure~\ref{fig:egg_cake_evaluation} compares \mymodel{} with $\pi_{0.5}$ and OpenWAM.
Vela achieves the highest success rate on all four subtasks, with a mean subtask success rate of 67.5\%, compared with 37.5\% for $\pi_{0.5}$ and 40.0\% for OpenWAM. The largest margin occurs during ingredient transfer, while the advantage persists for pan grasping and final plating. These results provide physical-robot evidence that \mymodel{}'s trajectory-space policy remains effective beyond simulation, particularly
during long-horizon sequencing, bimanual coordination, tool use, and contact-rich manipulation.
We provide additional real-world results on potato shredding in Appendix~\ref{app:additional_real_world}.
More visualizations of the real-world experiments are provided on the project page.

\subsection{Ablation Studies}\label{sec:ablation_study}

We ablate the core design of \mymodel{}. We examine whether trajectory-space learning offers benefits beyond post-hoc spline fitting, how motion-dependent horizons affect a fixed control-point budget, and whether the choice of geometry supervision is critical. These ablations distinguish the role of the learned representation from its temporal allocation and loss design.

\begin{table}
    \centering
    \vspace{-5pt}
    \caption{Ablation of post-hoc trajectory fitting (Post-hoc Traj. Fit.) on LIBERO-X. Post-hoc Traj. Fit. spline fits and resamples the predicted point-wise action chunk, whereas \mymodel{} predicts trajectory parameters directly. Max. Steps denotes the maximum number of actions executed before replanning.}
    \label{tab:ablation_trajectory}
    \resizebox{0.95\linewidth}{!}{
    \setlength{\tabcolsep}{5.5pt}
    \renewcommand{\arraystretch}{1.10}
    \begin{tabular}{lcccccccc}
        \toprule
        \multicolumn{3}{c}{\textbf{Configuration}}
        & \multicolumn{6}{c}{\textbf{Success Rate (\%)}} \\
        \cmidrule(lr){1-3}
        \cmidrule(lr){4-9}
        \textbf{Method}
        & \textbf{Post-hoc Traj. Fit.}
        & \textbf{Max. Steps}
        & \textbf{Level 1}
        & \textbf{Level 2}
        & \textbf{Level 3}
        & \textbf{Level 4}
        & \textbf{Level 5}
        & \textbf{Avg.} \\
        \midrule

        & \xmark
        & 10
        & 65.2
        & 53.2
        & 36.0
        & 24.1
        & 18.0
        & 39.3 \\

        $\pi_{0.5}$
        & \cmark
        & 10
        & 61.7
        & 50.5
        & 37.0
        & 24.7
        & 20.4
        & 38.9 \\

        & \cmark
        & 25
        & 62.6
        & 52.2
        & 38.8
        & 24.7
        & 21.9
        & 40.0 \\

        $\pi_{0.5}$-\mymodel & N/A & 25
        & \underline{66.1} & \underline{53.5} & \underline{40.7} & \underline{27.3} & \underline{22.5} & \underline{42.0} \\

        \midrule

        \textbf{\mymodel}
        & N/A
        & 25
        & \textbf{67.2}
        & \textbf{55.9}
        & \textbf{44.9}
        & \textbf{31.5}
        & \textbf{27.1}
        & \textbf{45.3} \\

        \bottomrule
    \end{tabular}}
\end{table}

\noindent\textbf{\ding{172} Can Post-hoc Trajectory Fitting Recover the Gains?}
Since trajectory fitting can provide a continuous path for action resampling, one may ask whether the benefit of \mymodel{} can simply come from a smoother trajectory.
To test this, we compare two ways of introducing trajectory structure from the
same $\pi_{0.5}$ base checkpoint. The post-hoc baseline retains action chunk prediction during post-training, then fits and resamples a spline at inference. In contrast, $\pi_{0.5}$-Vela learns to predict trajectory parameters during post-training.
Alongside the official 10-step configuration, we retrain a
$\pi_{0.5}$ baseline to support execution of up to 25 actions per
prediction, the largest per-observation execution steps supported by the official evaluator.
As shown in Tab.~\ref{tab:ablation_trajectory}, post-hoc fitting alone does not recover the gains of \mymodel{}'s modeling under all execution settings. This advantage is observed without our trajectory-space pretraining stage, supporting the value of learning trajectory parameters itself rather than imposing spline structure only after point-wise action prediction.

\noindent\textbf{\ding{173} Does Motion-Dependent Horizon Adaptation Matter?}
We next isolate the contribution of motion-dependent horizon adaptation from \mymodel{}'s modeling.
We compare a globally fixed horizon with motion-dependent adaptive horizon while retaining the same spline representation and control-point budget.
As shown in Tab.~\ref{tab:ablation_horizon}, \mymodel{}'s modeling provides gains even under a fixed temporal support, and horizon adaptation further raises performance at every level.
This result supports allowing the same trajectory capacity to cover different temporal spans, rather than imposing a uniform temporal allocation on every predicted motion.

\begin{table}[t]
    \centering
    \vspace{-5pt}
    \caption{
    Effect of adaptive temporal support on LIBERO-X. We compare \mymodel{} with globally fixed temporal support (Fixed H.) against the full model with adaptive temporal support (Adap. H.).
    }
    \label{tab:ablation_horizon}
    \resizebox{0.85\linewidth}{!}{
    \setlength{\tabcolsep}{7pt}
    \renewcommand{\arraystretch}{1.10}
    \begin{tabular}{lcccccc}
        \toprule
        \textbf{Method} & \textbf{Level 1} & \textbf{Level 2} & \textbf{Level 3} & \textbf{Level 4} & \textbf{Level 5} & \textbf{Avg.} \\
        \midrule

        $\pi_{0.5}$ & 65.2 & 53.2 & 36.0 & 24.1 & 18.0 & 39.3 \\

        \mymodel{} w/ Fixed H. & \underline{66.4} & \underline{54.3} & \underline{42.8} & \underline{28.1} & \underline{24.4} & \underline{43.2} \\

        \mymodel{} w/ Adap. H. & \textbf{67.2} & \textbf{55.9} & \textbf{44.9} & \textbf{31.5} & \textbf{27.1} & \textbf{45.3} \\
        \bottomrule
    \end{tabular}
    }
\end{table}

\noindent\textbf{\ding{174} Control Point vs. Decoded Trajectory Supervision.}\label{sec:objective_ablation}
CP-FM and DT-FM supervise the same control-point flow residual with different metrics: CP-FM uses Euclidean distance in parameter space, whereas DT-FM measures the residual after spline decoding. We compare CP-FM, an equal-weight combination of CP-FM and DT-FM, and DT-FM alone.
All three objectives yield effective policies, with DT-FM achieving
the highest average SR in Tab.~\ref{tab:ablation_loss}. The modest differences suggest that effective trajectory-space learning is \emph{not} contingent on a particular loss composition; decoder-aware weighting provides a useful refinement rather than a prerequisite.

\begin{table}
    \centering
    \caption{
    Ablation of trajectory geometry objectives on LIBERO-X.
    }
    \label{tab:ablation_loss}
    \resizebox{0.8\linewidth}{!}{
    \setlength{\tabcolsep}{7pt}
    \renewcommand{\arraystretch}{1.10}
    \begin{tabular}{cccccccc}
        \toprule
        \multicolumn{2}{c}{\textbf{Objectives}}
        & \multicolumn{6}{c}{\textbf{Success Rate (\%)}} \\
        \cmidrule(lr){1-2}
        \cmidrule(lr){3-8}
        \textbf{CP-FM}
        & \textbf{DT-FM}
        & \textbf{Level 1}
        & \textbf{Level 2}
        & \textbf{Level 3}
        & \textbf{Level 4}
        & \textbf{Level 5}
        & \textbf{Avg.} \\
        \midrule

        \cmark &   & 65.2 & 54.7 & 43.1 & \underline{31.3} & 25.8 & 44.0 \\

        \cmark & \cmark & \underline{65.7} & \textbf{56.0} & \underline{44.0} & \underline{31.3} & \textbf{27.6} & \underline{44.9} \\

          & \cmark & \textbf{67.2} & \underline{55.9} & \textbf{44.9} & \textbf{31.5} & \underline{27.1} & \textbf{45.3} \\

        \bottomrule
    \end{tabular}
    }
\end{table}

\section{Conclusion}

We introduced \mymodel{}, a trajectory-space vision-language-action foundation model designed to address the limitations of fixed-rate action chunks. It predicts cubic B-spline control points together with a motion-dependent temporal span, using a shared action interface and scalable target construction to represent heterogeneous robot motion as continuous trajectories. Our experiments show that native trajectory-space learning improves action modeling, while post-hoc spline fitting is insufficient. Motion-dependent temporal support lets a fixed representation adapt its resolution to different motions, and decoder-aware supervision and large-scale heterogeneous pretraining further improve transfer across precision, mobile, and long-horizon tasks.
Experiments on two real-world manipulation tasks further show that trajectory-space action modeling transfers beyond simulation to multi-stage bimanual manipulation involving tool use and sustained physical contact.
Overall, these results support continuous trajectories as a promising action representation for future embodied foundation models.

\bibliography{iclr2027_conference}
\bibliographystyle{iclr2027_conference}

\newpage
\appendix
\section*{Appendix}
\section{Additional Methodology Details}

\subsection{Target Construction}\label{app:target_construction}

For each sampled transition, the allocator selects the longest candidate horizon that satisfies both the continuous-coordinate and binary or near-deterministic channel (\eg event channel) fitting criteria.
If no candidate satisfies both criteria, we return the shortest candidate and its fitted control points as a fallback.
This fallback keeps target construction defined, but does not imply that the adequacy criteria are satisfied.
The selected spline fit and horizon form the joint supervision target. Algorithm~\ref{alg:vela_targets} makes this selection explicit.

\begin{algorithm}[htbp]
\caption{Motion-Dependent Horizon Adaptation}
\label{alg:vela_targets}
\begin{algorithmic}[1]

\Require Candidate horizons $\mathcal H=\{H_1>\cdots>H_K\}$, demonstrated action
$\mathbf A_{H_{1}}\in\mathbb R^{H_{1}\times d_a}$,
noise scales $\{\hat{\sigma}_j\}_{j\in\mathcal J}$,
tolerances $\kappa$ and $\eta$

\Ensure Adaptive trajectory target $(\mathbf C^\star,H^\star)$

\State Globally compute fixed spline basis and pseudoinverse pairs
$\{\mathbf B_H,\mathbf B_H^\dagger\}_{H\in\mathcal H}$

\For{$H\in\mathcal H$} \Comment{longest to shortest}

    \State
    $\mathbf A_H
    \gets
    \mathbf A_{H_{1}}[0{:}H,:]$

    \State
    $\mathbf C_H
    \gets
    \mathbf B_H^\dagger \mathbf A_H$

    \State
    $\mathbf R_H
    \gets
    \mathbf A_H-\mathbf B_H\mathbf C_H$

    \State
    $\displaystyle
    \psi^2(H)
    \gets
    \frac{1}{|\mathcal J|(H-N)}
    \sum_{j\in\mathcal J}
    \frac{\|\mathbf R_H^{(:,j)}\|_2^2}
         {\hat{\sigma}_j^2}$
    \Comment{continuous coordinates}

    \State
    $\displaystyle
    e_H
    \gets
    \wedge_{\substack{j\notin\mathcal J}}[
    \|\mathbf R_H^{(:,j)}\|_2/\sqrt H
    \le \eta]$
    \Comment{event coordinates}

    \If{$\psi(H)\le\kappa$ \textbf{and} $e_H$}
        \State \Return $(\mathbf C_H,H)$
    \EndIf

\EndFor

\State \Return $(\mathbf C_{H_K},H_K)$
\Comment{fallback to the shortest horizon}

\end{algorithmic}
\end{algorithm}

\noindent\textbf{Computation.}
Although Algorithm~\ref{alg:vela_targets} is written as a descending scan to make the target-selection rule explicit, all candidate horizons can be evaluated jointly in implementation. The spline basis matrices and their pseudo-inverses depend only on the configured candidate horizons and the fixed spline configuration, and are therefore precomputed once and reused throughout training.
In practice, we evaluate all candidates in a single batched pass and select the largest admissible horizon by a masked reduction.

\subsection{Objective Details}\label{app:objective}
Sec.~\ref{sec:mid_training} defines DT-FM by applying the spline decoder to the control point flow residual. Here we make explicit its relation to CP-FM. Since $\hat{A}_H=B_HC$, the control point flow error $\mathbf E_C$ induces the decoded action trajectory error $\mathbf B_{H^\star}\mathbf E_C$ at the target horizon $H^\star$. Thus, DT-FM can be viewed as applying flow supervision in the decoded action space rather than directly in the control point space,
\begin{equation}
    \ell_{\mathrm{DT}}
    =
    \frac{
        \|\mathbf B_{H^\star}\mathbf E_C\|_F^2
    }{
        H^\star d_a
    }
    =
    \frac{1}{d_a}
    \operatorname{tr}\!\left[
        (\mathbf E_C)^\top
        \frac{
            \mathbf B_{H^\star}^\top\mathbf B_{H^\star}
        }{
            H^\star
        }
        (\mathbf E_C)
    \right].
    \label{eq:app_decoder_weighting}
\end{equation}
Thus, DT-FM differs from isotropic control point supervision only
through the positive-semidefinite metric induced by spline decoding.
For the spline configurations used in our experiments,
$\mathbf B_{H^\star}$ has full column rank, such that
$\mathbf B_{H^\star}^\top\mathbf B_{H^\star}$ is positive definite.
Consequently,
\[
    \ell_{\mathrm{DT}}=0
    \quad\Longleftrightarrow\quad
    \mathbf E_C=\mathbf 0
    \quad\Longleftrightarrow\quad
    \ell_{\mathrm{CP}}=0.
\]
This establishes that both losses detect the same nonzero control-point flow errors while weighting their directions differently.
It does not, by itself, establish identical minimizers of the expected training objectives: the decoder metric depends on the target horizon, which may remain random conditional on the noisy flow input.
Accordingly, our comparison treats DT-FM as a decoder-induced reweighting and evaluates its effect empirically.

\section{Additional Implementation Details}\label{app:additional_imple_details}
\subsection{Trajectory-Space Pretraining Details}\label{app:pretraining_details}
This section describes the large-scale trajectory-space pretraining of \mymodel{} from pretrained PaliGemma VLM weights. $\pi_{0.5}$-Vela does not undergo this stage: it is adapted from the $\pi_{0.5}$ base checkpoint using the benchmark-specific post-training procedures below.

\mymodel\ is optimized with AdamW
($\beta_1=0.9$, $\beta_2=0.95$, $\epsilon=10^{-8}$, weight decay $10^{-10}$)
and global gradient clipping at $1.0$. The learning rate is linearly warmed up to $5\times10^{-5}$ over the first 10k optimization steps and kept constant thereafter. We apply a $0.5\times$ learning-rate multiplier to the PaliGemma backbone, corresponding to $2.5\times10^{-5}$, while the action expert and output projection use the full learning rate. Training proceeds for 300k optimization steps.

Normalization statistics are computed independently for each data source and action dimension using a quantile transform that maps the 1st and 99th percentiles to $[-1,1]$, while unavailable action dimensions are excluded from supervision through validity masks.

For adaptive target construction, only the two gripper coordinates are treated as event channels; all remaining valid action coordinates use the noise-calibrated adequacy criterion. We set the global adequacy threshold to $\kappa=2.5$ and the event-channel reconstruction tolerance to $\eta=0.05$.

Training is distributed over 64 NVIDIA H200 GPUs using FSDP with gradient checkpointing, with 24 samples per GPU, yielding a
global batch size of 1,536.
Transformer computation runs under bfloat16 autocast, while B-spline target construction, flow interpolation, and loss computation run in fp32 for numerical stability.

\subsection{Benchmark Fine-tuning Details And Evaluation}\label{app:benchmark_finetuning_details}

\noindent\textbf{Initialization and post-training.}
For each benchmark, $\pi_{0.5}$, $\pi_{0.5}$-Vela, and \mymodel{} use the same training split and optimization hyperparameters. The $\pi_{0.5}$ baseline and $\pi_{0.5}$-Vela start from the same $\pi_{0.5}$ base checkpoint and are post-trained with action chunk and trajectory-space supervision, respectively. \mymodel{} starts from the trajectory-space pretrained checkpoint described in Appendix~\ref{app:additional_imple_details}.
The optimization settings and execution configurations are specified separately below.

\noindent\textbf{LIBERO-X.}
LIBERO-X~\citep{wang2026libero} provides 2,520 human-teleoperated training demonstrations spanning
600 tasks and 100 scenes. The demonstrations are collected at 20\, Hz, with third-person and wrist-camera observations and Cartesian end-effector actions together with gripper commands. Following the official $\pi_{0.5}$ configuration, visual inputs are resized to $224\times224$.

For benchmark-specific fine-tuning, we follow the official $\pi_{0.5}$
full-parameter fine-tuning recipe. Fine-tuning is performed for 30k optimization steps with a global batch size of 256. We use AdamW
($\beta_1=0.9$, $\beta_2=0.95$, $\epsilon=10^{-8}$, weight decay $10^{-10}$)
with global gradient clipping at $1.0$. The learning rate is linearly warmed up to $5\times10^{-5}$ over the first 10k steps and kept constant thereafter, and an exponential moving average with decay $0.999$ is maintained throughout fine-tuning.

Evaluation follows the official five-level protocol. Level 1 introduces local spatial perturbations, Level 2 expands the spatial variation, Level 3 reconstructs scene topology, Level 4 further varies visual attributes and introduces unseen or confounding objects, and Level 5 additionally reformulates instructions while preserving their task semantics. The perturbations are cumulative, such that each successive level introduces
additional sources of distribution shift. Each task is evaluated over 10 rollouts with the benchmark-defined initial-state configuration and rollout perturbations. Following the official setting, the task-specific time limit is set to $1.1\times$ the average human completion time. We use the benchmark implementation without modifying environment initialization, perturbation ranges, or success conditions. The LIBERO-X evaluator executes at most 25 actions from each prediction. For a predicted action sequence containing $h_t$ actions, the executed length is $\min(h_t,25)$.

The official $\pi_{0.5}$ benchmark configuration uses 10-step execution. For the 25-step post-hoc comparison in Tab.~\ref{tab:ablation_trajectory}, we retrain a separate $\pi_{0.5}$ baseline from the same base checkpoint with its horizon configuration adjusted to support execution of up to 25 actions,
while retaining the other optimization hyperparameters. Spline fitting is then applied to this model's predicted action chunks at inference.

\noindent\textbf{EBench.}
We follow the official $\pi_{0.5}$ fine-tuning recipe released with EBench~\citep{gao2026ebench}. The benchmark training set contains 6,600 demonstrations (91.4 hours) across all 26 tasks, including 2,800 human-teleoperated trajectories for the 7 tabletop dexterous-and-precise tasks and 3,800 motion-planned trajectories for the 10 mobile pick-and-place and 9 mobile long-horizon tasks.

All parameters are fine-tuned from the corresponding initialization for 200k optimization steps with a global batch size of 128. We use AdamW ($\beta_1=0.9$, $\beta_2=0.95$, $\epsilon=10^{-8}$, weight decay $10^{-10}$) with global gradient clipping at $1.0$. The learning rate is linearly warmed up for 10k steps to $5\times10^{-5}$ and subsequently cosine-decayed to $5\times10^{-6}$ over the full 200k-step schedule. An exponential moving average with decay $0.999$ is maintained throughout fine-tuning.

The Test set contains 510 episodes spanning the four generalization dimensions. The EBench evaluator executes at most 30 actions from each
prediction. For a predicted action sequence containing $h_t$ actions, the executed length is $\min(h_t,30)$.

\section{Further Analysis on Experiments}\label{app:further_exp}

\begin{figure}
    \centering
    \includegraphics[width=\linewidth]{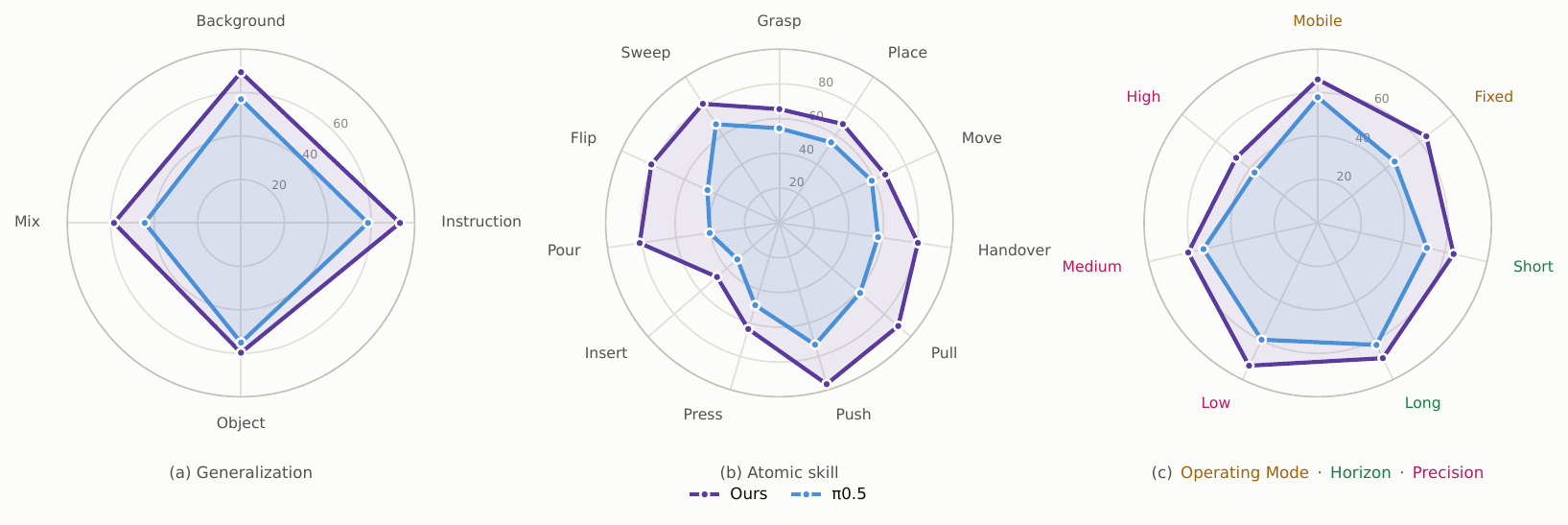}
    \caption{Detailed capability profile of \mymodel{} and $\pi_{0.5}$ on EBench. We report the official task progress Score (\%) across (a) the four generalization conditions, (b) atomic skills, and (c) operating mode, temporal horizon, and precision. \mymodel{} consistently improves over $\pi_{0.5}$ across the displayed categories}
    \label{fig:ebench_radar_score}
\end{figure}

\begin{figure}
    \centering
    \includegraphics[width=\linewidth]{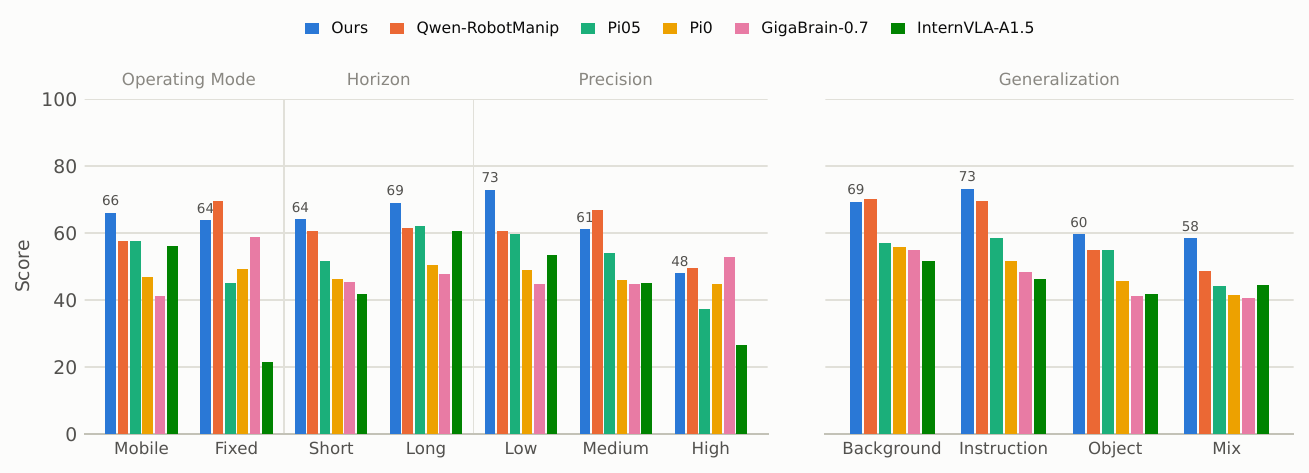}
    \caption{Fine-grained capability breakdown on EBench. We compare \mymodel{} with representative baselines using the official task progress Score (\%) across operating mode (Mobile / Fixed-base), temporal horizon (Short / Long), precision (Low / Medium / High), and four generalization conditions (Background / Instruction / Object / Mix).}
    \label{fig:ebench_bars_score}
\end{figure}

\noindent\textbf{Additional results on EBench.}
The main text summarizes EBench performance by its three manipulation task families. Here, we provide a finer-grained view by exploiting the benchmark's task-level and capability annotations.
Table~\ref{tab:task_success} reports task-level success rates, while Figures~~\ref{fig:ebench_radar_score} and~\ref{fig:ebench_bars_score} reorganize the evaluation along operating mode, temporal horizon, precision, generalization, and atomic skill.
These views address complementary questions: task-level SR measures complete task success, whereas the grouped Score profiles describe progress across broader evaluation regimes.

\begin{table*}
\centering
\caption{Task-level results on EBench. Success rate (SR, \%) for each of the 26 tasks, grouped into dexterous-and-precise tabletop,
mobile pick-and-place (PnP), and mobile long-horizon tasks. The final row reports benchmark-wide performance.}
\label{tab:task_success}

\begingroup
\footnotesize
\setlength{\tabcolsep}{3pt}
\renewcommand{\arraystretch}{1.12}

\begin{tabular*}{\textwidth}{
  @{\extracolsep{\fill}}p{0.27\textwidth}>{\columncolor{gray!15}}r*{8}{r}@{}
}
\toprule
Task
& \scriptsize\textbf{Ours}
& \scriptsize\shortstack{Qwen-\\RobotManip}
& \scriptsize$\pi_{0.5}$
& \scriptsize\shortstack{InternVLA-\\A1.5}
& \scriptsize$\pi_0$
& \scriptsize\shortstack{GigaBrain-\\0.7}
& \scriptsize\shortstack{Cosmos3-\\Edge}
& \scriptsize FastWAM
& \scriptsize X-VLA \\
\midrule

\multicolumn{10}{@{}l}{\textit{TableTop}} \\[2pt]
collect coffee beans
& 15 & 5 & 30 & 0 & 20 & \textbf{50} & 0 & 0 & 0 \\
flip cup collect cookies
& 60 & \textbf{90} & 25 & 30 & 30 & 40 & 30 & 0 & 5 \\
frame against pen holder
& 65 & \textbf{95} & 45 & 0 & 25 & 70 & 20 & 0 & 30 \\
install gear
& 25 & \textbf{50} & 10 & 15 & 30 & 20 & 15 & 15 & 5 \\
peg in hole
& 15 & 10 & 10 & 0 & 15 & 15 & \textbf{30} & 0 & 5 \\
put glass in glassbox
& \textbf{90} & \textbf{90} & 65 & 0 & 85 & 55 & 25 & 5 & 15 \\
tighten nut
& \textbf{40} & 10 & 20 & \textbf{40} & 25 & 15 & 10 & 0 & 0 \\

\midrule
\multicolumn{10}{@{}l}{\textit{PnP}} \\[2pt]
apple from shelf
& 45 & 65 & \textbf{75} & 35 & 40 & 55 & 15 & 40 & 70 \\
apple to fruit bowl
& 50 & 50 & 25 & 20 & 10 & 20 & 5 & \textbf{70} & 50 \\
bookmark on book
& \textbf{55} & 40 & \textbf{55} & 45 & 25 & 30 & 25 & 30 & 40 \\
bowl to plate
& 80 & 75 & \textbf{85} & 60 & 60 & 65 & 65 & 75 & 80 \\
perfume to cosmetics rack
& \textbf{80} & 55 & 60 & 45 & 25 & 25 & 60 & 50 & 35 \\
remote to holder
& \textbf{80} & 65 & 60 & 60 & 70 & 50 & 75 & 50 & 35 \\
salt to spice rack
& \textbf{100} & 70 & 70 & 90 & 80 & 75 & 80 & 80 & 65 \\
soap to dish
& 85 & 85 & \textbf{100} & 95 & 95 & 70 & 85 & 90 & 70 \\
teacup to saucer teapot to tray
& \textbf{50} & 45 & 30 & 30 & 10 & 20 & 0 & 10 & 10 \\
utensils to holder
& 5 & 15 & 5 & 20 & 0 & 10 & 5 & 0 & \textbf{45} \\

\midrule
\multicolumn{10}{@{}l}{\textit{LongHorizon}} \\[2pt]
bottle
& 0 & 0 & 0 & 0 & 0 & 0 & 0 & 0 & 0 \\
detergent
& 90 & 85 & \textbf{100} & 95 & 75 & 95 & 95 & 35 & 0 \\
dish
& 5 & 20 & \textbf{25} & 20 & 0 & 5 & 0 & 0 & 0 \\
dishwasher
& \textbf{90} & 15 & 65 & \textbf{90} & \textbf{90} & 0 & 15 & 70 & 50 \\
fruit
& 20 & \textbf{45} & 10 & 5 & 0 & 0 & 0 & 0 & 0 \\
make sandwich
& 73 & 67 & 67 & 33 & 20 & \textbf{80} & 27 & 47 & 7 \\
microwave
& 53 & 33 & 40 & 60 & 47 & 0 & \textbf{80} & 0 & 0 \\
pen
& 0 & \textbf{5} & 0 & 0 & 0 & 0 & 0 & 0 & 0 \\
shop
& \textbf{20} & 0 & 0 & 0 & 0 & 0 & 0 & 0 & 0 \\

\midrule
Overall
& \textbf{49.7} & 45.6 & 41.4 & 34.2 & 33.7 & 33.3 & 29.3 & 25.6 & 23.7 \\
\bottomrule
\end{tabular*}

\par\smallskip

\endgroup
\end{table*}

Figure~\ref{fig:ebench_radar_score} further isolates the comparison with $\pi_{0.5}$. Across all three profiles, \mymodel{} consistently exceeds $\pi_{0.5}$: the improvement holds across all four generalization conditions, atomic skill categories, and operating mode, horizon, and precision. The resulting profile indicates that the overall gain is broad-based rather than concentrated in a particular task family or capability category.

Figure~\ref{fig:ebench_bars_score} places this breadth in the context of other policies: \mymodel{} performs strongly across operating modes, task horizons, and generalization conditions, without leading every precision category.
Together, the task-level and grouped results support the use of one
trajectory representation across varied manipulation demands, rather than an advantage confined to a small task subset.

\begin{figure}
    \centering
    \includegraphics[width=0.9\linewidth]{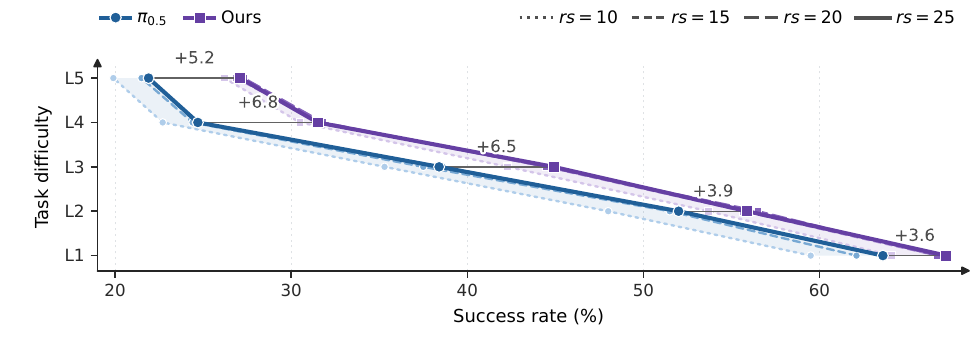}
    \caption{Robustness to replanning interval on LIBERO-X. We vary the maximum number of executed actions before replanning for both $\pi_{0.5}$ and \mymodel{}.
 \label{fig:replan_steps}}
\end{figure}

\noindent\textbf{Robustness Across Execution Caps.} Temporal support is learned as part of \mymodel{}'s action representation, whereas the execution cap limits how many decoded actions are consumed before replanning. To assess whether \mymodel{} 's advantage depends on a particular execution schedule, we evaluate both $\pi_{0.5}$ and \mymodel{} with caps of 10, 15, 20, and 25 actions. For \mymodel{}, a cap $r$ gives an executed length of $\min(h_t,r)$ without fixing the predicted trajectory span. As shown in Fig.~\ref{fig:replan_steps}, \mymodel{} consistently outperforms $\pi_{0.5}$ across the entire range of execution caps and exhibits less variation with this deployment choice. These results show that the benefit of \mymodel{}'s modeling is not tied to a narrowly tuned execution schedule, but persists across different execution cap settings.

\section{Additional Real-World Evaluation Details}\label{app:additional_real_world}
\noindent\textbf{Robot platform and evaluation protocol.}
We conduct all real-world experiments on the same wheeled dual-arm robot introduced in Sec.~\ref{sec:exp_setup}. The platform has two 7-DoF arms and three RGB cameras: one mounted on the head and one on each wrist. We evaluate Vela in two manipulation settings: the four-stage egg-cake cooking task reported in the main paper and an additional four-stage potato-shredding task.

\noindent\textbf{Real-World Evaluation on Potato Shredding.}
We additionally evaluate \mymodel{} on the \emph{shredding potato} task using the same wheeled dual-arm robot.
The task requires coordinated bimanual tool manipulation and repeated physical contact between the potato and grater.
The task is decomposed into four subtasks as follows:
\begin{enumerate}
    \item \textbf{Prepare tools.}
    Pick up a potato with the left gripper and the grater with the right
    gripper.

    \item \textbf{Contact and shred.}
    Adjust the potato against the grater, then repeatedly grate it so the shreds fall into the water.

    \item \textbf{Return tools.}
    Return the potato and grater to their designated locations.

    \item \textbf{Remove shreds.}
    Use the strainer to lift the potato shreds out of the water and transfer them into the bowl.
\end{enumerate}

\begin{figure}[H]
    \centering
    \includegraphics[width=\linewidth]{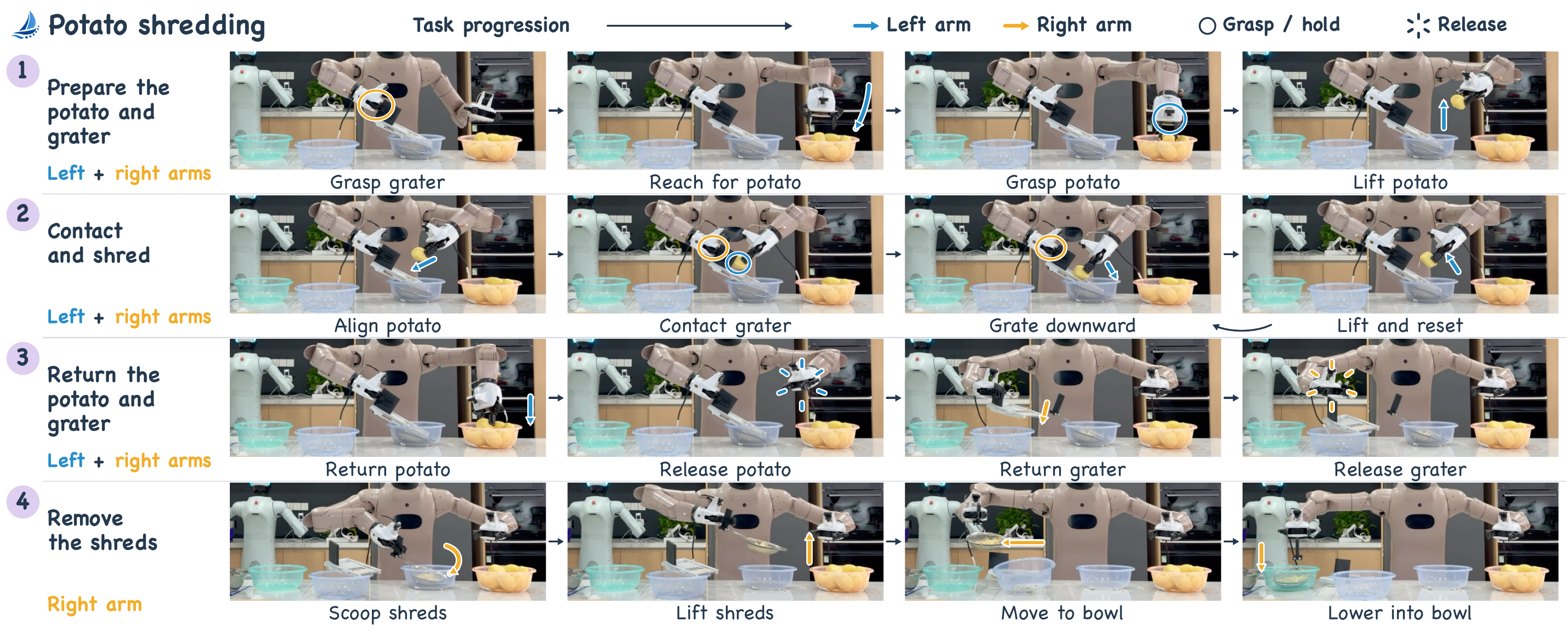}
    \caption{\mymodel{} executing \emph{shredding potato}, which spans four subtasks (top to bottom).}
    \label{fig:demo_potato_shredding}
\end{figure}

\begin{figure}
    \centering
    \includegraphics[width=\linewidth]{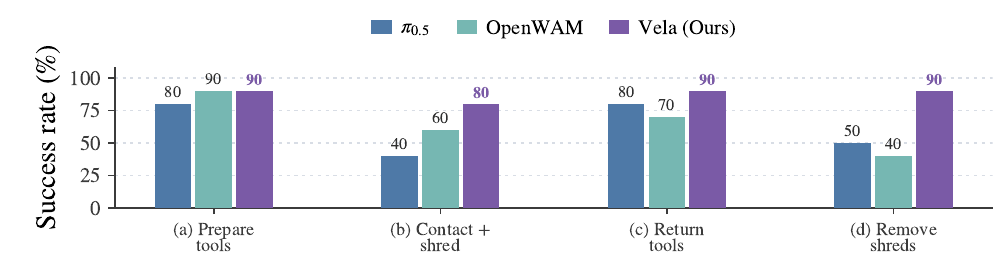}
    \caption{Additional real-world potato-shredding evaluation. Each bar reports the success rate over ten trials for one of the four ordered subtasks.}
    \label{fig:real_world_potato_shredding}
\end{figure}

For $\pi_{0.5}$ and OpenWAM, we report the corresponding results under the same four-subtask protocol.
\mymodel{} achieves a mean subtask success rate of $87.5\%$, compared with $62.5\%$ for $\pi_{0.5}$ and $65.0\%$ for OpenWAM, while matching or exceeding both baselines on every stage. The gains are most evident in stages involving sustained physical contact and coordinated tool manipulation, providing complementary real-world evidence to the egg-cake experiment in the main paper.

\section{Limitations}
Vela provides a starting point for exploring how trajectory-space action modeling scales with model capacity and training data. The present study uses a model at the scale of $\pi_{0.5}$, leaving larger vision-language backbones and more extensive pretraining as natural extensions. Studying these settings would help characterize how the benefits of trajectory-space modeling evolve with model size and data coverage.
Expanding the action expert offers a complementary direction within the same continuous trajectory interface. Building on the base-model capacity used here, larger or specialized action experts could be explored for fine contact dynamics and long-horizon bimanual coordination. Evaluating these designs would clarify how additional action-expert capacity affects the generation of trajectories for demanding manipulation tasks.
A further direction is to adapt representation capacity to motion complexity. The current policy adapts the temporal horizon while keeping the number of control points fixed. Allowing both quantities to vary could allocate compact representations to smooth motions and additional detail to more complex trajectories. This extension would require variable-length prediction and corresponding training objectives to balance representation efficiency with motion fidelity.

\end{document}

%% file: math_commands.tex
\usepackage{amsmath,amsfonts,bm}

\def\eqref#1{equation~\ref{#1}}
\def\1{\bm{1}}

\DeclareMathAlphabet{\mathsfit}{\encodingdefault}{\sfdefault}{m}{sl}
\SetMathAlphabet{\mathsfit}{bold}{\encodingdefault}{\sfdefault}{bx}{n}